\documentclass[runningheads]{llncs}
\usepackage{graphicx}
\graphicspath{ {./images/} }
\usepackage{hyperref}
\usepackage{natbib}
\begin{document}
\title{Why You Should Try the Real Data for the Scene Text Recognition}
%
%
\author{Loginov Vladimir \inst{1}\orcidID{0000-0001-6952-0037} }
\authorrunning{V. Loginov}
%
\institute{Intel Corporation \email{vladimir.loginov@intel.com}}
\maketitle              
\begin{abstract}
  Recent works in the text recognition area have pushed forward the recognition results
  to the new horizons. But for a long time a lack of large human-labeled natural text recognition
  datasets has been forcing researchers to use synthetic data for training
  text recognition models. Even though synthetic datasets are very large
  (MJSynth and SynthText, two most famous synthetic datasets, have several
  million images each), their diversity could be insufficient, compared to
  natural datasets like ICDAR and others. Fortunately, the recently released
  text recognition annotation for OpenImages V5 dataset has comparable with
  synthetic dataset number of instances and more diverse examples. We have
  used this annotation with a Text Recognition head architecture from the Yet
  Another Mask Text Spotter and got comparable to the SOTA results. On some
  datasets we have even outperformed previous SOTA models. In this paper we also
  introduce a text recognition model.
  The model's code is available.
  \footnote{\url{https://github.com/openvinotoolkit/training_extensions}}

\end{abstract}
\section{Introduction}
The text recognition plays an important role in people's lives.
It can be used for digitalizing old documents and helping blind people read.
Also, the text recognition could be utilized in a variety of automation algorithms;
one of the most common use-cases is the license plate recognition.

Since the release of the synthetic datasets (MJSynth\cite{Jaderberg14c},
SynthText\cite{Gupta16}), a common approach to the model training was the following:
use synthetic data for training and real data for evaluation. This was caused by a
lack of labeled large text recognition datasets. From the very first deep learning text
recognition architectures such as \cite{crnn} to the modern works like \cite{cstr},
\cite{cui2021representation} researchers have followed this pattern. Even though
real datasets have train splits, their size is significantly less than the size of
synthetic datasets.

Due to the lack of real data for training, previous works aimed mainly at tuning an architecture
of the model rather than using different data. \cite{baek2019STRcomparisons} suggested the
idea of differentiating a text recognition model into 4 stages: transformation, feature
extraction, sequence modelling and prediction. Many of the existing text recognition model
architectures fit into this scheme. We follow the same approach: in our work we
use TPS \cite{tpsstn} for transformation, ResNeXt \cite{ResNeXt} for feature
extraction, convolutional encoder-recurrent decoder from YAMTS \cite{krylov2021open}
with 2D attention mechanism the sequence modelling and the prediction.

The rest of the paper is organized as follows:
\begin{itemize}
  \item In the Section \ref{sec:rel_works} we briefly discuss other works in this field.
  \item In the Section \ref{sec:model_arch} we describe in details the suggested model architecture.
  \item In the Section \ref{sec:experiments} we show the results of our experiments.
  \item In the Section \ref{sec:conclusion} we make conclusions and do some call to action for other
        researchers.
\end{itemize}

Our main contributions are as follows:
\begin{itemize}
  \item Developing the idea of using real data together with synthetic data for training text recognition models.
  \item Suggesting text recognition model which performs on par with other models and
        even outperforms current state-of-the-art approaches on some datasets.
\end{itemize}

\section{Related Works}
\label{sec:rel_works}

MJSynth\cite{Jaderberg14c}, SynthText\cite{Gupta16} are two large (each has several
million images with text) text recognition datasets. These two datasets contain images
generated in the certain pipeline. In MJSynth, image generation process is the following:
first, certain font is picked and text in this font is rendered; second, using the image from the
previous step, borders and shadows are colored and composed together; third, projective
distortion is applied to the image; and the last step is natural image blending. See
\cite{Jaderberg14c} for details. SynthText's pipeline takes into account depth of the image
regions and uses segmentation network to predict suitable positions to put the text.
See the corresponding publication \cite{Gupta16} for details.

These two works play an important role in text recognition area, and their contribution
could hardly be overestimated. However, even though images from MJSynth and SynthText look like
real, in some cases it is clear that the specific image is manually generated. Moreover, real
world images are sometimes so various and rich in texture, curvature, color and other visual
features that it is hard to generate them using a set of predefined rules. \textbf{Thus, there
  is a problem}: there are datasets used for training with great number of examples, but
these examples are not diverse enough; and there are testing datasets
with more complex, real images on which the model is evaluated. This causes a misalignment
between training and testing samples. The problem is also known as domain shift
\cite{michieli2020adversarial, Zhang_2019_CVPR}. Model fits to the training images and, without
augmentations, shows poorer performance on the testing ones. If a researcher decides
to use augmentation to deal with the overfitting problem, model spends its capacity to
learn augmented training samples which are still different from testing. This results in
model performance drop on the testing dataset.

\cite{baek2021STRfewerlabels} was one of the first works which suggested the idea of using
real datasets for training text recognition models, but in unsupervised manner, since
large annotated datasets were not available then. Recently released text spotting
annotation for OpenImagesV5 \cite{krylov2021open} and TextOCR \cite{textocr} made
training text recognition models on real data in a supervised manner possible.

ASTER \cite{aster} is one of the works which utilizes Thin Plate Spline transformation for
automatic image rectification. Experiments from \cite{aster} show that this technique
helps to boost the recognition accuracy.

ResNeXt \cite{ResNeXt} and ResNet-like \cite{resnet} backbones are widely used as feature
extractors for many text recognition models (\cite{aster, cstr, qiao2020seed,
  baek2019STRcomparisons}). We have experimented extensively with the configuration of the
backbone and found that standard configuration without the last stage pretrained on the
ImageNet\cite{imagenet} performs the best.

Attention mechanism \cite{bahdanau2016neural} was first introduced as a novel technique for
the machine translation. Recently, the attention mechanism have become a standard for text
recognition models.

Alongside with annotation for text spotting we also use the text recognition
head from \cite{krylov2021open}. It performs well for text spotting task and
our experiments show that it also performs well for text recognition task.

\section{Model Architecture}
\label{sec:model_arch}
\subsection{Overall Architecture}
\begin{figure}[t]
  \centering
  \includegraphics[width=\textwidth]{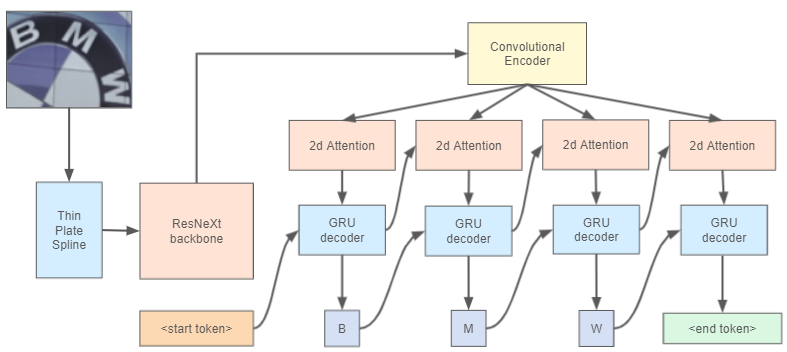}
  \caption{Proposed text recognition model architecture. }
  \label{figure:model}
\end{figure}

We follow the same approach as in \cite{baek2019STRcomparisons} but do not divide the last two
stages. Thus, the text recognition head consists of sequence modelling and final prediction.
We utilize TPS \cite{tpsstn} as in ASTER \cite{aster} for image rectification, since this stage
has become one of the standard building blocks for modern text recognition models.
We use ResNeXt \cite{ResNeXt} for feature extraction due to its good performance/accuracy
trade-off. In our work we have utilized the same text recognition head as
in \cite{krylov2021open}. You can see the full model in the Figure~\ref{figure:model}.

\subsection{Thin Plate Spline}
Thin Plate Spline (TPS) \cite{tpsstn} was first proposed for image recognition in the
rectification task. Later, this module was utilized in some text recognition models
\cite{aster, qiao2020seed}.

The spatial transformer mechanism is split into three parts. In order of computation,
first a localization network takes the image and generates parameters for spatial transformation.
Using the predicted parameters, a sampling grid is generated, which is a set of points
where the input map should be sampled to produce the transformed output. This is done by
the grid generator. Finally, the sampler generates output image using the grid and input image.

Thin plate spline is designed in such way that it does not require letter level annotation.
Thus, it can be utilized as is in many text recognition architectures and its parameters
can be learned in an end-to-end manner.

\subsection{Backbone}
Since the release of the ResNet \cite{resnet} this network has become popular for many
computer vision tasks due to its simplicity of learning and capacity.
During the past years, ResNet-like networks showed their effectiveness and were
used in many other text recognition models: ASTER \cite{aster}, CSTR \cite{cstr},
SEED \cite{qiao2020seed}, \cite{baek2019STRcomparisons}.
The ResNeXt-101 consists of 5 stages. We use ResNeXt-101
without the last stage. We find this configuration to have a good accuracy/performance trade-off.
In our model we utilize pretrained on ImageNet \cite{imagenet} weights as
initial for the ResNeXt-101.

\subsection{Text Recognition Head}
Our network heavily utilizes the text recognition head from \cite{krylov2021open} with minor
changes. First, we increase the number of channels to 1024 in the convolutional encoder since
this is the exact number of the output channels from the ResNext-101 (without the last stage).
Convolutional encoder of the head also produces 1024 output channels. Second, for the same
reason, we have changed the size of the feature map to $3 \times 12$. For the rest of the
parameters, we use the same architecture, including the depth of the encoder part and the type
of the recurrent decoder (GRU \cite{gru}).

\section{Experiments}
\label{sec:experiments}
\subsection{Datasets}

\subsubsection{Training Datasets}
There are two large synthetic datasets:
\begin{itemize}
  \item MJSynth is a synthetically generated text recognition dataset \cite{Jaderberg14c}.
        It has about 9M images.
  \item SynthText is also a synthetic dataset, but it has an annotation for text detection
        and text recognition. It consists of about 800k images with approximately 8M word instances.
\end{itemize}
Recently there were released two large natural text recognition datasets: TextOCR \cite{textocr}
and OpenImages V5 text spotting \cite{krylov2021open}.

\begin{itemize}
  \item TextOCR has about 900k instances of annotated words.
  \item OpenImages V5 text spotting dataset has about 2M text instances.
\end{itemize}

In our work we have used a combination of synthetic and real data: MJSynth and OpenImages V5 text spotting.
We did not use any specific technique to balance samples from different datasets in the batch.

\subsubsection{Testing Datasets}
In the text recognition task, datasets could be divided into two groups: Regular and Irregular datasets.
In the regular datasets, like ICDAR03, ICDAR13, IIIT5k, SVT, text is mainly horizontally oriented without
curvatures and perspective distortions. Irregular datasets, like ICDAR15, SVT-P, CUTE, mostly consist
of curved texts, which could be with some perspective distortions. Images from irregular datasets are
usually more challenging for text recognition models to make the prediction.

We perform evaluation of our model on the testing parts of the following datasets:

\textbf{Regular datasets:}
\begin{itemize}
  \item ICDAR2003 \cite{icdar03} consists of 1156 images for training and 1110 images for evaluation.
        A common practice for text recognition models is to ignore all images which contain non-alphanumeric
        symbols and are less than 3 symbols in length. This results in 867 images for evaluation.
  \item ICDAR2013 \cite{icdar13} is inherited from ICDAR2003. Its train and test split have
        848 and 1095 images, respectively. Ignoring non-alphanumeric words results in the dataset of
        size 1015.
  \item IIIT5k \cite{iiit5k} is collected from Google image search. It has 2k images for training
        and 3k for testing.
  \item SVT \cite{svt} has 257 images for training and 647 images for evaluation. Its name stands
        for Street View Text and consists of outdoor street images.
\end{itemize}

\textbf{Irregular datasets:}
\begin{itemize}
  \item ICDAR2015 \cite{icdar15} was created for the ICDAR 2015 Robust Reading competition
        and contains 4468 images for training and 2077 images for evaluation. All evaluation subset
        is used for the evaluation, even though it consists of some examples with non-alphanumeric
        instances. Unlike the previous datasets, this has examples with irregular text.
  \item SVT-P \cite{svtp} is also collected from Google Street View and has many examples with
        perspective distortions. Dataset size is 645.
\end{itemize}

Both training and evaluation of the model were performed in case-insensitive mode. The vocabulary
of our experiments consists of 10 digits, 26 letters and 4 special symbols: start token,
end token, pad token and unknown token. Pad token is used in training phase for numeric
representations of the texts to be equal length and further concatenate them into one batch.
Unknown token is used to represent symbol that is not in the vocabulary (e.g., question mark).
Thus, the final stage of our model predicts every character among 40 classes.
We do not use any lexicon or beam search to boost the recognition accuracy.

\subsection{Image Preprocessing}
Image preprocessing and augmentation plays an important role in many deep learning algorithms.
We resize all input images to $64 \times 256$ fixed resolution. TPS preprocessing of our network
produces image of fixed size $48 \times 192$. We also use color jittering (from torchvision
\cite{torchvision}) and Gaussian blurring (from opencv \cite{opencv_library}) as augmentations.
In the end of the augmentation pipeline we convert all images to grayscale and scale
images to be in [0;1] interval.

\subsection{Loss Function}
Standard Negative Loglikelihood loss function is used as cost function. We use Adam
\cite{kingma2017adam} optimizer with a constant learning rate 1e-4. Other parameters are
default for PyTorch's Adam optimizer, including zero weight decay.
The model was trained for 15 epochs.

\subsection{Environment}
All experiments were performed on a server with 8 NVIDIA\textregistered\ Tesla P100 GPUs with 16 GB VRAM. We
use PyTorch 1.8.1 as our training framework. Batch size was set to 48.

\subsection{Evaluation Results}
We compare our model with current state-of-the art approaches in the Table~\ref{table:evaluation}.
Standard text recognition accuracy is used as a metric of quality.

\begin{table}[ht]
  \centering
  \caption{Evaluation results and comparison with other models.
    Results in bold are the best in every dataset.}
  \begin{tabular}{|l|l|l|l|l|l|l|l|}
    \hline
    Model                                     & IC03          & IC13          & IC15          & IIIT5K        & SVT           & SVT-P         \\
    \hline
    CSTR \cite{cstr}                          & 94.8          & 93.2          & 81.6          & 90.1          & 93.7          & 85.0          \\
    ASTER \cite{aster}                        & 94.8          & 93.2          & 81.6          & 90.1          & 93.7          & 85.0          \\
    SEED \cite{qiao2020seed}                  & -             & 92.8          & 80.0          & 93.8          & 89.6          & 81.4          \\
    RCEED \cite{cui2021representation}        & -             & 94.7          & \textbf{82.2} & \textbf{94.9} & 91.8          & 83.6          \\
    SATRN \cite{lee2020satrn}                 & 96.7          & 94.1          & 79.0          & 92.8          & 91.3          & 86.5          \\
    Baek et al. \cite{baek2019STRcomparisons} & 94.4          & 92.3          & 71.8          & 87.9          & 87.9          & 79.2          \\
    \textbf{Our model}                        & \textbf{97.1} & \textbf{96.8} & 80.2          & 93.5          & \textbf{94.7} & \textbf{89.9} \\ [1ex] 
    \hline
  \end{tabular}

  \label{table:evaluation}
\end{table}

\subsection{Ablation Study}
\subsubsection{Results Using Different Train Datasets}

During our research, we have experimented with different datasets used in training. You can see
experiments' results in the Table~\ref{table:dataset_comparison}. All experiments were performed
under the same conditions: ResNeXt-101 backbone, YAMTS text recognition head, case-insensitive
training, case-insensitive evaluation. We did not use TPS module for this experiment.
\begin{table}[ht]
  \centering
  \caption{Comparison of the recognition accuracy using different training datasets.}
  \begin{tabular}{|l|l|l|l|l|l|l|l|}
    \hline
    Dataset                                  & IC03          & IC13          & IC15          & IIIT5K      & SVT           & SVT-P         \\
    \hline
    MJSynth                                  & 92.7          & 88.9          & 64.2          & 84.2        & 86.2          & 78.1          \\
    MJSynth + SynthText                      & 95.2          & 93.5          & 69.3          & 87.5        & 88.1          & 82.9          \\
    \textbf{MJSynth} + \textbf{OpenImagesV5} & \textbf{95.6} & \textbf{95.2} & \textbf{75.3} & \textbf{92} & \textbf{91.8} & \textbf{87.1} \\
    OpenImages V5                            & 94.1          & 94.2          & 73.2          & 91.1        & 89.0          & 82.9          \\
    \hline
  \end{tabular}

  \label{table:dataset_comparison}
\end{table}

It is clear from the table, that having only synthetic data is insufficient to train a
robust text recognition model. It is also interesting that using only real data drops
recognition accuracy a bit. Possible explanation for this fact is that since the training
examples are more difficult, the model should be trained on a longer schedule
(possibly with lower learning rate) like in case of strong augmentations.

\subsubsection{Influence of Case Sensitive Learning}

As it was stated in the previous section, our first models were trained in case-sensitive
mode while evaluation was performed in case-insensitive mode, like in \cite{lee2020satrn}.
This was caused by the overfitting problem, and case-sensitive training could be thought as a
special augmentation type. Since the OpenImagesV5 dataset is more challenging than
synthetic dataset, augmentation could be reduced, allowing model to spend its capacity
on learning features of the target dataset, and not augmentations. We have compared
two training setups, the only difference is case mode (and, thus, number of output
classes). The results are reported in the Table~\ref{table:case_mode}.

\begin{table}[ht]
  \centering
  \caption{Comparison of the recognition accuracy using different case mode.}
  \begin{tabular}{|l|l|l|l|l|l|l|l|}
    \hline
    Case Mode        & IC03 & IC13 & IC15 & IIIT5K & SVT  & SVT-P \\
    \hline
    Case-Sensitive   & 95.6 & 95.2 & 75.3 & 92.0   & 91.8 & 87.1  \\
    Case-Insensitive & 97.0 & 96.5 & 78.2 & 93.5   & 93.5 & 91.6  \\
    \hline
  \end{tabular}

  \label{table:case_mode}
\end{table}

Note, these two experiments are performed without TPS module.

\subsection{Experiments with ViTSTR}

Transformer models have become very popular during the past months for many deep learning
problems, and scene text recognition is not an exception. We have tried to train ViTSTR
\cite{vitstr} using the OpenImagesV5 annotation. The results can be seen in the table~
\ref{table:vit}. We have trained the models in the same framework, but, for an honest comparison,
we have disabled filtering non-alphanumeric symbols during the test. Numbers near the name of the
datasets mean portion of the dataset in the batch. Thus, 0.5 ST + 0.5 MJ means balanced batch
containing examples from both MJSynth and SynthText. Baseline model is ViTSTR-Tiny, both models
were trained using random data augmentation from the source framework.

\begin{table}[ht]
  \centering
  \caption{Comparison of the recognition accuracy using different training datasets for ViTSTR-Tiny+Aug.}
  \begin{tabular}{|l|l|l|l|l|l|l|}
    \hline
    Training Dataset                  & IC03 & IC13 & IC15 & IIIT5K & SVT  & SVT-P \\
    \hline
    0.5MJ + 0.5 ST (baseline)         & 93.7 & 89.8 & 66.6 & 85.1   & 85.9 & 78.5  \\
    0.4MJ + 0.2 OpenImagesV5 + 0.4 ST & 95.6 & 94.2 & 77.0 & 91.7   & 90.7 & 86.7  \\
    \hline
  \end{tabular}

  \label{table:vit}
\end{table}

The intention of this experiment was not to train the best model, but to show
that the model actually have higher capacity and could perform better on the
validation datasets, if trained on the real data. The more difficult is the
dataset (e.g. IC15 and SVT-P), the higher is the accuracy gain. It is clear from the table
that even though large augmentation was applied to train the
baseline model, not full capacity of the model was utilized.

\section{Conclusion}
\label{sec:conclusion}
In this paper we have shown  that training on natural text recognition datasets
could be more efficient than on synthetically generated images. Our experiments have
proven the fact that training with more challenging data and less augmentations is more
efficient in terms of accuracy on the testing dataset. We have also
introduced a text recognition model which achieves state-of-the-art results
on some common text recognition benchmarks. We encourage other researchers to
try to use the real data to train their models, because as we have shown in the
Section~\ref{sec:experiments}, this can help to boost the recognition accuracy,
especially, for the attention-based and transformer models.

\bibliographystyle{splncs04nat}
\bibliography{bibliography}

\begin{thebibliography}{30}
\providecommand{\natexlab}[1]{#1}
\providecommand{\url}[1]{\texttt{#1}}
\providecommand{\urlprefix}{URL }
\expandafter\ifx\csname urlstyle\endcsname\relax
  \providecommand{\doi}[1]{doi:\discretionary{}{}{}#1}\else
  \providecommand{\doi}{doi:\discretionary{}{}{}\begingroup
  \urlstyle{rm}\Url}\fi

\bibitem[{Atienza(2021)}]{vitstr}
Atienza, R.: Vision transformer for fast and efficient scene text recognition.
  CoRR \textbf{abs/2105.08582} (2021),
  \urlprefix\url{https://arxiv.org/abs/2105.08582}

\bibitem[{Baek et~al.(2019)Baek, Kim, Lee, Park, Han, Yun, Oh, and
  Lee}]{baek2019STRcomparisons}
Baek, J., Kim, G., Lee, J., Park, S., Han, D., Yun, S., Oh, S.J., Lee, H.: What
  is wrong with scene text recognition model comparisons? dataset and model
  analysis. In: International Conference on Computer Vision (ICCV) (2019)

\bibitem[{Baek et~al.(2021)Baek, Matsui, and Aizawa}]{baek2021STRfewerlabels}
Baek, J., Matsui, Y., Aizawa, K.: What if we only use real datasets for scene
  text recognition? toward scene text recognition with fewer labels. In:
  IEEE/CVF Conference on Computer Vision and Pattern Recognition (CVPR) (2021)

\bibitem[{Bahdanau et~al.(2016)Bahdanau, Cho, and Bengio}]{bahdanau2016neural}
Bahdanau, D., Cho, K., Bengio, Y.: Neural machine translation by jointly
  learning to align and translate (2016)

\bibitem[{Bradski(2000)}]{opencv_library}
Bradski, G.: {The OpenCV Library}. Dr. Dobb's Journal of Software Tools  (2000)

\bibitem[{Cai et~al.(2021)Cai, Sun, and Xiong}]{cstr}
Cai, H., Sun, J., Xiong, Y.: {CSTR:} {A} classification perspective on scene
  text recognition. CoRR \textbf{abs/2102.10884} (2021),
  \urlprefix\url{https://arxiv.org/abs/2102.10884}

\bibitem[{Cho et~al.(2014)Cho, van Merri{\"e}nboer, Gulcehre, Bahdanau,
  Bougares, Schwenk, and Bengio}]{gru}
Cho, K., van Merri{\"e}nboer, B., Gulcehre, C., Bahdanau, D., Bougares, F.,
  Schwenk, H., Bengio, Y.: Learning phrase representations using {RNN}
  encoder{--}decoder for statistical machine translation. In: Proceedings of
  the 2014 Conference on Empirical Methods in Natural Language Processing
  ({EMNLP}), pp. 1724--1734, Association for Computational Linguistics, Doha,
  Qatar (Oct 2014), \doi{10.3115/v1/D14-1179},
  \urlprefix\url{https://www.aclweb.org/anthology/D14-1179}

\bibitem[{Cui et~al.(2021)Cui, Wang, Zhang, and Wang}]{cui2021representation}
Cui, M., Wang, W., Zhang, J., Wang, L.: Representation and correlation enhanced
  encoder-decoder framework for scene text recognition (2021)

\bibitem[{Gupta et~al.(2016)Gupta, Vedaldi, and Zisserman}]{Gupta16}
Gupta, A., Vedaldi, A., Zisserman, A.: Synthetic data for text localisation in
  natural images. In: IEEE Conference on Computer Vision and Pattern
  Recognition (2016)

\bibitem[{He et~al.(2016)He, Zhang, Ren, and Sun}]{resnet}
He, K., Zhang, X., Ren, S., Sun, J.: Deep residual learning for image
  recognition. In: 2016 IEEE Conference on Computer Vision and Pattern
  Recognition (CVPR), pp. 770--778 (2016), \doi{10.1109/CVPR.2016.90}

\bibitem[{Jaderberg et~al.(2014)Jaderberg, Simonyan, Vedaldi, and
  Zisserman}]{Jaderberg14c}
Jaderberg, M., Simonyan, K., Vedaldi, A., Zisserman, A.: Synthetic data and
  artificial neural networks for natural scene text recognition. In: Workshop
  on Deep Learning, NIPS (2014)

\bibitem[{Jaderberg et~al.(2015)Jaderberg, Simonyan, Zisserman, and
  kavukcuoglu}]{tpsstn}
Jaderberg, M., Simonyan, K., Zisserman, A., kavukcuoglu, k.: Spatial
  transformer networks. In: Cortes, C., Lawrence, N., Lee, D., Sugiyama, M.,
  Garnett, R. (eds.) Advances in Neural Information Processing Systems,
  vol.~28, Curran Associates, Inc. (2015),
  \urlprefix\url{https://proceedings.neurips.cc/paper/2015/file/33ceb07bf4eeb3da587e268d663aba1a-Paper.pdf}

\bibitem[{Karatzas et~al.(2013)Karatzas, Gomez-Bigorda, Nicolaou, Ghosh,
  Bagdanov, Iwamura, Matas, Neumann, Chandrasekhar, Lu, Shafait, Uchida, and
  Valveny}]{icdar13}
Karatzas, D., Gomez-Bigorda, L., Nicolaou, A., Ghosh, S., Bagdanov, A.,
  Iwamura, M., Matas, J., Neumann, L., Chandrasekhar, V.R., Lu, S., Shafait,
  F., Uchida, S., Valveny, E.: {ICDAR 2013 robust reading competition}. pp.
  1484--1493 (08 2013), \doi{10.1109/ICDAR.2013.221}

\bibitem[{Karatzas et~al.(2015)Karatzas, Gomez-Bigorda, Nicolaou, Ghosh,
  Bagdanov, Iwamura, Matas, Neumann, Chandrasekhar, Lu, Shafait, Uchida, and
  Valveny}]{icdar15}
Karatzas, D., Gomez-Bigorda, L., Nicolaou, A., Ghosh, S., Bagdanov, A.,
  Iwamura, M., Matas, J., Neumann, L., Chandrasekhar, V.R., Lu, S., Shafait,
  F., Uchida, S., Valveny, E.: {ICDAR 2015 competition on Robust Reading}. In:
  {2015 13th International Conference on Document Analysis and Recognition
  (ICDAR)}, pp. 1156--1160 (2015), \doi{10.1109/ICDAR.2015.7333942}

\bibitem[{Kingma and Ba(2017)}]{kingma2017adam}
Kingma, D.P., Ba, J.: Adam: A method for stochastic optimization (2017)

\bibitem[{Krylov et~al.(2021)Krylov, Nosov, and Sovrasov}]{krylov2021open}
Krylov, I., Nosov, S., Sovrasov, V.: Open images v5 text annotation and yet
  another mask text spotter (2021)

\bibitem[{Lee et~al.(2020)Lee, Park, Baek, Oh, Kim, and Lee}]{lee2020satrn}
Lee, J., Park, S., Baek, J., Oh, S.J., Kim, S., Lee, H.: On recognizing texts
  of arbitrary shapes with 2d self-attention. In: Proceedings of the IEEE
  Conference on Computer Vision and Pattern Recognition (CVPR) Workshop on Text
  and Documents in the Deep Learning Era WTDDLE (2020)

\bibitem[{Lucas et~al.(2003)Lucas, Panaretos, Sosa, Tang, Wong, and
  Young}]{icdar03}
Lucas, S., Panaretos, A., Sosa, L., Tang, A., Wong, S., Young, R.: {ICDAR 2003
  robust reading competitions}. vol. 682-687, pp. 682-- 687 (09 2003), ISBN
  0-7695-1960-1, \doi{10.1109/ICDAR.2003.1227749}

\bibitem[{Marcel and Rodriguez(2010)}]{torchvision}
Marcel, S., Rodriguez, Y.: Torchvision the machine-vision package of torch. In:
  Proceedings of the 18th ACM International Conference on Multimedia, p.
  1485–1488, MM '10, Association for Computing Machinery, New York, NY, USA
  (2010), ISBN 9781605589336, \doi{10.1145/1873951.1874254},
  \urlprefix\url{https://doi.org/10.1145/1873951.1874254}

\bibitem[{Michieli et~al.(2020)Michieli, Biasetton, Agresti, and
  Zanuttigh}]{michieli2020adversarial}
Michieli, U., Biasetton, M., Agresti, G., Zanuttigh, P.: Adversarial learning
  and self-teaching techniques for domain adaptation in semantic segmentation
  (2020)

\bibitem[{Mishra et~al.(2012)Mishra, Alahari, and Jawahar}]{iiit5k}
Mishra, A., Alahari, K., Jawahar, C.V.: Scene text recognition using higher
  order language priors. In: BMVC (2012)

\bibitem[{Phan et~al.(2013)Phan, Shivakumara, Tian, and Tan}]{svtp}
Phan, T.Q., Shivakumara, P., Tian, S., Tan, C.L.: Recognizing text with
  perspective distortion in natural scenes. In: ICCV, pp. 569--576 (2013)

\bibitem[{Qiao et~al.(2020)Qiao, Zhou, Yang, Zhou, and Wang}]{qiao2020seed}
Qiao, Z., Zhou, Y., Yang, D., Zhou, Y., Wang, W.: Seed: Semantics enhanced
  encoder-decoder framework for scene text recognition (2020)

\bibitem[{Russakovsky et~al.(2014)Russakovsky, Deng, Su, Krause, Satheesh, Ma,
  Huang, Karpathy, Khosla, Bernstein, Berg, and Li}]{imagenet}
Russakovsky, O., Deng, J., Su, H., Krause, J., Satheesh, S., Ma, S., Huang, Z.,
  Karpathy, A., Khosla, A., Bernstein, M.S., Berg, A.C., Li, F.: {ImageNet
  Large Scale Visual Recognition Challenge}. CoRR \textbf{abs/1409.0575}
  (2014), \urlprefix\url{http://arxiv.org/abs/1409.0575}

\bibitem[{Shi et~al.(2015)Shi, Bai, and Yao}]{crnn}
Shi, B., Bai, X., Yao, C.: An end-to-end trainable neural network for
  image-based sequence recognition and its application to scene text
  recognition. CoRR \textbf{abs/1507.05717} (2015),
  \urlprefix\url{http://arxiv.org/abs/1507.05717}

\bibitem[{Shi et~al.(2019)Shi, Yang, Wang, Lyu, Yao, and Bai}]{aster}
Shi, B., Yang, M., Wang, X., Lyu, P., Yao, C., Bai, X.: Aster: An attentional
  scene text recognizer with flexible rectification. IEEE Transactions on
  Pattern Analysis and Machine Intelligence \textbf{41}(9), 2035--2048 (2019),
  \doi{10.1109/TPAMI.2018.2848939}

\bibitem[{Singh et~al.(2021)Singh, Pang, Toh, Huang, Galuba, and
  Hassner}]{textocr}
Singh, A., Pang, G., Toh, M., Huang, J., Galuba, W., Hassner, T.: {TextOCR:
  Towards large-scale end-to-end reasoning for arbitrary-shaped scene text}.
  CoRR \textbf{abs/2105.05486} (2021),
  \urlprefix\url{https://arxiv.org/abs/2105.05486}

\bibitem[{Wang et~al.(2011)Wang, Babenko, and Belongie}]{svt}
Wang, K., Babenko, B., Belongie, S.: End-to-end scene text recognition. In:
  ICCV, pp. 1457--1464, IEEE (2011)

\bibitem[{Xie et~al.(2016)Xie, Girshick, Doll{\'{a}}r, Tu, and He}]{ResNeXt}
Xie, S., Girshick, R.B., Doll{\'{a}}r, P., Tu, Z., He, K.: Aggregated residual
  transformations for deep neural networks. CoRR \textbf{abs/1611.05431}
  (2016), \urlprefix\url{http://arxiv.org/abs/1611.05431}

\bibitem[{Zhang et~al.(2019)Zhang, Nie, Liu, Xu, Zhang, and
  Shen}]{Zhang_2019_CVPR}
Zhang, Y., Nie, S., Liu, W., Xu, X., Zhang, D., Shen, H.T.:
  Sequence-to-sequence domain adaptation network for robust text image
  recognition. In: Proceedings of the IEEE/CVF Conference on Computer Vision
  and Pattern Recognition (CVPR) (June 2019)

\end{thebibliography}

\end{document}